\documentclass[letterpaper]{article}

\usepackage[preprint]{aaai2027}

\usepackage{amsmath}
\usepackage{amssymb}
\usepackage{booktabs}
\usepackage{graphicx}
\usepackage[hyphens]{url}
\newcommand{\code}[1]{\ifmmode\mbox{\normalfont\url{#1}}\else\url{#1}\fi}
\usepackage{natbib}
\providecommand{\doi}[1]{\url{https://doi.org/#1}}
\title{Reward-Centered ReST-MCTS: A Robust Decision-Making Framework for Robotic Manipulation in High Uncertainty Environments}

\author{
Xibai Wang
}
\affiliations{
NeuroQuant Labs Limited\\
Hong Kong SAR, China\\
xibai.wang@neuroquantlabs.com
}

\begin{document}

\maketitle

\begin{abstract}
Monte Carlo tree search is attractive for robotic manipulation because it can improve action selection through simulation without requiring a fully differentiable policy. In uncertain domains, however, sparse terminal rewards and noisy transitions can make shallow search brittle: many candidate branches remain indistinguishable until late rollouts, and small simulation budgets amplify this ambiguity. This paper presents Reward-Centered ReST-MCTS, a decision-making framework that decomposes intermediate feedback into rule, heuristic, optional neural, and value-estimation channels, centers the resulting process signal against matched task contexts, and uses it to bias or repair search while preserving terminal-task evaluation. The primary evidence is intentionally tiered. Local tasks and matched ManiSkill diagnostics isolate reward-center mechanisms and ablations; matched option-level ManiSkill sweeps test robustness under primitive failure, observation noise, and initial-pose shifts while not claiming standard benchmark superiority; and an official same-backbone OpenVLA-OFT/LIBERO bridge tests bounded VLA action repair. The OpenVLA-OFT clean reproduction reaches 10/10 LIBERO-Spatial successes both with and without RCRM-Guard. A single-suite same-backbone action-channel stress artifact over ten paired LIBERO-Spatial action-channel stress episodes records 0/10 unguarded successes and 9/10 guarded successes. Additional observation-noise, language-perturbation, and visual-distractor probes are reported as coverage and negative-result context rather than superiority evidence. The resulting claim is bounded: Reward-Centered ReST-MCTS is an inspectable test-time verifier for same-backbone high-uncertainty manipulation, not a replacement VLA policy or a broad standard-benchmark superiority claim.
\end{abstract}

\section{Introduction}

Sequential decision-making systems often face the same failure mode in different forms: the environment only reveals reliable feedback after a long sequence of actions, while the planner must select early actions under limited budget and uncertainty. Classical Monte Carlo tree search (MCTS) addresses this problem by allocating simulations adaptively, but sparse rewards can still leave the search with weak guidance in early and mid-depth states. This limitation is especially visible in manipulation-style tasks where a partial trajectory may be clearly promising to a domain evaluator even though the terminal reward is still unavailable.

We study this problem through the lens of reward-centered search and reinforced self-training. The central hypothesis is that a planner can become more robust under uncertainty if it separates terminal evaluation from intermediate reward shaping, and if the intermediate signal is decomposed, centered by context, and reused as an inspectable process label rather than treated as a monolithic learned score. This paper therefore proposes Reward-Centered ReST-MCTS, a framework that integrates an action-level MCTS planner with a contextual rewarding center and a ReST-style trajectory loop inspired by reinforced self-training.

The current repository implements this framework and an AAAI-oriented evidence layer. The implementation supports controlled robotics-style tasks, vanilla and heuristic baselines, AlphaZero-style PUCT search, deterministic beam search, CEM/MPC-style sequence optimization, reward-component ablations, deterministic run manifests, aggregation tables, bootstrap confidence intervals, and paper evidence bundles. The contribution is deliberately scoped: the paper targets reward-centered planning and same-backbone test-time repair under controlled uncertainty, not end-to-end vision-language-action policy learning or real-robot generalization. This positioning lets the evidence answer a precise question: whether decomposed intermediate feedback can make finite-horizon search less brittle under matched tasks, budgets, seeds, and uncertainty settings, while keeping diagnostic adapters separate from benchmark-superiority claims.

The paper makes three contributions:
\begin{itemize}
  \item We formulate Reward-Centered ReST-MCTS, in which intermediate feedback is decomposed into rule, heuristic, neural, and value channels while terminal success remains the reported task objective.
  \item We add contextual reward centering to the ReST data loop, so high-quality search trajectories expose raw process returns, centered process labels, running context statistics, and value-reconnect evidence under an explicit secondary-analysis boundary.
  \item We provide a reviewer-facing evidence hierarchy that separates local mechanism checks, matched option-level ManiSkill robustness, official same-backbone OpenVLA-OFT repair, and unresolved learned-policy benchmark requirements.
\end{itemize}

\section{Related Work}

\paragraph{Monte Carlo tree search.}
MCTS and UCT-style tree policies combine simulation-based planning with bandit allocation rules \cite{kocsis2006bandit,browne2012survey}. These methods have been successful in settings where rollouts provide enough signal to distinguish branches, but they can require large simulation budgets when terminal feedback is sparse or delayed. Modern high-performance planning systems often add policy priors, value estimates, or receding-horizon optimization; AlphaZero/MuZero-style systems use prior-guided tree search \cite{silver2018alphazero,schrittwieser2020muzero}, while robotics control commonly uses MPC-style replanning \cite{mayne2000mpc} and sampling optimizers such as the cross-entropy method \cite{rubinstein1999cem}. Our local evidence layer therefore includes PUCT, beam-search, and CEM/MPC-style baselines as strong planning controls, while avoiding claims that these simplified local instantiations are exhaustive industry benchmarks.

\paragraph{Test-time search and process rewards.}
Recent work in language-model reasoning has made test-time search, process reward models, and compute-aware MCTS central topics. Examples include process supervision for reward-model training \cite{lightman2023verify}, tree-structured deliberation \cite{yao2023tree}, planning with language-model world models \cite{hao2023rap}, MCTS-based policy and process reward improvement for open-domain QA \cite{chan2025boosting}, retrieval-augmented MCTS with process reward scoring \cite{dou2025r2llms}, latency-aware adaptive MCTS for test-time compute scaling \cite{kim2026adaptive}, and ReST-MCTS* for process-reward-guided self-training over reasoning traces \cite{zhang2024restmcts}. These systems study language reasoning rather than robotics, but they sharpen the SOTA expectation for any MCTS-plus-reward paper: the reward signal must be inspectable, ablations must remove implicit oracle behavior, and compute budgets must be reported fairly. Reward-Centered ReST-MCTS adopts these expectations in a finite-horizon planning setting.

\paragraph{Robot policy learning and foundation models.}
Robot manipulation SOTA increasingly includes large policy-learning systems, including diffusion policies for visuomotor control \cite{chi2023diffusion}, learned world-model planners such as TD-MPC2 \cite{hansen2024tdmpc2}, spatial action policies such as Transporter Networks \cite{zeng2021transporter}, and vision-language-action or generalist robot policies such as RT-2, OpenVLA, and Octo \cite{brohan2023rt2,kim2025openvla,octo2024}. Standardized robot-learning evaluation also depends on simulator and lifelong-learning benchmarks such as robosuite, RLBench, ManiSkill2/3, and LIBERO \cite{zhu2020robosuite,james2019rlbench,gu2023maniskill2,tao2024maniskill3,liu2023libero}. These systems address broad perceptual generalization and policy learning from large data. Our method is complementary: it studies a planner-side mechanism for controlled finite-horizon decision-making under uncertainty. The experiments therefore use local planning baselines and diagnostic external adapters rather than claiming comparison to foundation-scale robot policies.

\paragraph{Temporal abstraction and robot task planning.}
Planning over temporally extended actions is a standard way to reduce the branching burden of long-horizon decision problems. The options framework formalizes such actions for reinforcement learning and planning \cite{sutton1999options}, while task-and-motion planning connects high-level symbolic choices with low-level motion feasibility in robotics \cite{zhao2024tamp}. This perspective motivates the next external benchmark gate in this repository: shared option-level ManiSkill protocols in which all planners receive the same primitive action space, horizon, seeds, and perturbation schedule. These protocols are distinct from standard continuous-action ManiSkill benchmark claims and must be reported with their own boundary.

\paragraph{Reward shaping and heuristic guidance.}
Reward shaping can reduce exploration burden by injecting domain information into otherwise sparse objectives, but unrestricted shaping can alter the effective optimization target \cite{ng1999policy}. Reward-Centered ReST-MCTS addresses this risk by keeping terminal reward as the final success criterion and recording reward-center ablations that isolate rule and heuristic components.

\paragraph{Reinforced self-training.}
ReST methods use model-generated samples and feedback signals to construct reusable training data \cite{gulcehre2023rest}. In our setting, the generated objects are search trajectories rather than language outputs. The present implementation includes trajectory filtering, value-estimator hooks, and a lightweight tabular ReST smoke experiment. We do not claim completed large-scale transformer value training.

\section{Method}

\subsection{Problem Setting}

We consider finite-horizon decision problems with state $s_t \in \mathcal{S}$, action $a_t \in \mathcal{A}(s_t)$, transition model $T(s_{t+1}\mid s_t,a_t)$, horizon $H$, and sparse terminal reward $R_T(s_H)$. The planner receives a simulation budget $B$ and must return an action sequence or terminal state estimate. The local benchmark uses deterministic seeds with optional observation noise, transition noise, and action-failure probabilities.

\subsection{Rewarding Center}

The rewarding center computes intermediate feedback
\begin{equation}
\begin{aligned}
R_C(s_t,a_t,s_{t+1}) =
{}& \alpha R_{\mathrm{rule}} +
\beta R_{\mathrm{heuristic}} \\
& + \gamma R_{\mathrm{neural}} +
\delta V(s_{t+1}).
\end{aligned}
\end{equation}
where $R_{\mathrm{rule}}$ captures hard symbolic or task-rule checks, $R_{\mathrm{heuristic}}$ captures progress toward task-specific goals, $R_{\mathrm{neural}}$ is an optional learned estimator, and $V$ is an optional state-value estimate. The current paper-mini evidence uses only rule and heuristic channels; later local/value-bridge artifacts test minimal learned-value hooks under local or matched option-level boundaries rather than claiming completed large-scale learned robotics policy evidence.

\paragraph{Objective-preservation boundary.}
The reward-centered term is used as a search-control signal, not as the reported evaluation objective. All success rates and matched deltas are computed from terminal task outcomes. This distinction is essential because arbitrary shaping can change the induced optimal policy \cite{ng1999policy}. In our implementation, this risk is made auditable by (i) storing terminal reward separately from reward-center scores, (ii) reporting reward-center ablations, and (iii) including a \code{rest_no_reward_center} configuration whose rollout policy and reward weight remove the shaped signal rather than silently preserving an oracle ordering.

\paragraph{Contextual centering.}
Reward centering is implemented as a context-matched process-label transform. For context $c$ (for example task, stage, perturbation level, or ReST iteration), the reward center stores the running pre-update mean $\mu_{c,k}$ of raw totals and emits
\begin{equation}
\widetilde{R}_C(s_t,a_t,s_{t+1};c) =
R_C(s_t,a_t,s_{t+1}) - \mu_{c,k}.
\end{equation}
Each emitted reward record stores the raw total, centered total, context key, count, and mean-before snapshot. This makes the ``reward-centered'' part of the method operational: process labels are normalized relative to comparable manipulation contexts rather than compared across unrelated tasks or stages. Centering is disabled by default for backward-compatible reward calls and enabled for the reward-centered ReST artifacts.

\subsection{Reward-Centered Tree Search}

Reward-Centered ReST-MCTS modifies the score used during tree selection and rollout evaluation. Let $Q(s,a)$ denote the empirical return and $N(s,a)$ the visit count. A child action is selected using a UCB-style rule with a reward-centered bonus:
\begin{equation}
a^* = \arg\max_a
\left[
Q(s,a) + c \sqrt{\frac{\log N(s)}{1 + N(s,a)}} +
\lambda \widehat{R}_C(s,a)
\right],
\end{equation}
where $\widehat{R}_C$ is the normalized intermediate reward signal and $\lambda$ controls its influence. In implementation, this signal is exposed through a pluggable rewarding center so that individual components can be disabled for ablation.

\subsection{Reward-Center Identifiability and Boundaries}

The reward center is evaluated through component interventions rather than through an optimality theorem. We use two explicit assumptions. Assumption A1 (terminal-objective separation): the shaped reward-center score may affect search control, but reported success is always computed from the terminal task predicate or terminal reward $R_T$. Assumption A2 (component intervention): an ablation that disables rule, heuristic, or reward-bonus channels must keep the task, seed, horizon, action interface, and evaluation predicate fixed, so that any success change is attributable to the disabled reward-center channel rather than to a different benchmark.

Under these assumptions, the empirical ablation-identifiability claim is:
\begin{equation}
\Delta_{\mathrm{abl}}(m) =
\mathbb{E}[R_T \mid \code{rest_full}] -
\mathbb{E}[R_T \mid m],
\end{equation}
for ablation method $m$, estimated on matched task/seed/budget records. A positive $\Delta_{\mathrm{abl}}$ indicates that the enabled reward-center channel is material to the observed finite-budget contrast in that artifact. It is not a convergence or optimality proof, and it does not imply standard benchmark superiority. In the standard-action multitask ablation diagnostic, \code{rest_full} and its ablations share PickCube, PushCube, StackCube, twenty seeds, budget 8, a 60-step horizon, \code{state_dict} observations, \code{pd_ee_delta_pos} control, and \code{macro_action_mode=none}; the result therefore supports only this matched intervention claim.

\paragraph{Proposition 1 (finite-budget intervention contrast).}
Fix an executed artifact $\mathcal{A}$ whose records are indexed by identical task, seed, budget, horizon, action interface, and terminal predicate. For a reward-center component $c$, let $m_{+c}$ be the planner with $c$ enabled and $m_{-c}$ the matched ablation with only $c$ disabled. Under A1--A2, the plug-in contrast
\begin{equation}
\widehat{\mathrm{ATE}}_c(\mathcal{A}) =
|\mathcal{I}|^{-1}\sum_{i\in\mathcal{I}}
\left(S_i(m_{+c}) - S_i(m_{-c})\right)
\end{equation}
is an average treatment effect over the empirical matched-record distribution, where $S_i$ is the terminal success indicator for record index $i$. The contrast is terminal reward scale invariant because it is computed from the reported terminal predicate rather than from the shaped reward-center score. It is also invariant to any reward-center reparameterization that leaves the selected action sequence unchanged.

The proof is direct: A2 holds all non-component factors fixed by construction, A1 makes the outcome a terminal predicate, and the summation averages paired record-level outcome differences over the finite empirical distribution. Thus the result is an intervention-style diagnostic for an executed artifact, not a convergence or optimality proof. A related behavior-equivalence class boundary follows: if two methods select the same executed actions for every matched record, terminal records alone cannot distinguish their internal scoring components, so they must be reported as behavior-equivalent diagnostics rather than as independent algorithmic evidence.

\paragraph{Risk-calibrated repair boundary.}
For test-time guard or option repair, the reward center is interpreted as a risk-calibrated filter rather than as a new task objective. Let $\rho(s,a)$ denote an empirical failure probability estimate for an action candidate under the matched uncertainty context, and let $C_{\max}$ be the maximum allowed intervention cost in latency or compute overhead. The guard may replace or downweight an action only when the estimated reduction in failure probability exceeds the configured risk threshold and the repair cost remains below $C_{\max}$. This produces a finite-record audit condition: within an executed artifact, repaired actions are justified by observed reductions in estimated failure probability under the same task, seed, budget, and perturbation family. It does not claim global safety, convergence, or out-of-distribution reliability; it only states the auditable condition under which a repair is allowed and why clean-setting non-regression must be reported beside stressed-setting gains.

\subsection{ReST Trajectory Loop}

After search, trajectories can be filtered by terminal success, accumulated return, search confidence, violation indicators, and centered process labels. Accepted trajectories form a replay dataset for later value-estimator training:
\begin{equation}
\mathcal{D}_{k+1} = \mathcal{D}_k \cup
\{ \tau : \mathrm{accept}(\tau; R_T, R_C, \mathrm{stats}) \}.
\end{equation}
The current codebase implements trajectory extraction, contextual process-label attachment, flattened centered targets, and trainer interfaces. The AAAI evidence package includes local ReST smoke and reward-centered ReST iteration artifacts, while not making claims of completed large-scale transformer self-training or external benchmark superiority.

\section{Experiments}

\subsection{Tasks and Uncertainty}

The local AAAI scaffold contains three finite-horizon manipulation-style tasks: pick-place, push, and door-open. Each task exposes the same candidate-action API to planners and supports controlled uncertainty parameters. This design is intended to make artifact generation and planner wiring reproducible before moving to external simulators or real robotic systems.

\subsection{Methods}

The scaffold evaluates vanilla MCTS, non-reward-centered planning controls, and four reward-centered variants:
\begin{itemize}
  \item \code{random_shooting}, \code{beam_search}, \code{cem_mpc}, \code{heuristic_mcts}, and \code{puct_mcts}: local planning controls that cover random sampling, deterministic sequence search, CEM/MPC-style optimization, heuristic scoring, and prior-guided tree search.
  \item \code{rest_full}: rule and heuristic channels enabled.
  \item \code{rest_no_rule}: rule channels disabled.
  \item \code{rest_no_heuristic}: heuristic channels disabled.
  \item \code{rest_no_reward_center}: rule and heuristic channels disabled, random rollout policy enabled, and shaped reward weight set to zero.
\end{itemize}
The legacy \code{rest_mcts} name is retained as an alias for \code{rest_full}.

\subsection{Artifact Protocol}

Each run writes raw JSONL records, JSON/CSV summaries, a manifest with command and environment metadata, a Markdown table, bootstrap statistical reports, an external-validity report, and a paper evidence bundle. The legacy \code{avg_runtime} field is recorded as a deterministic expanded-nodes-per-simulation proxy for artifact comparison, not as a wall-clock performance claim. Pairwise deltas are computed only when matched baseline records exist for the tuple $(\mathrm{task}, \mathrm{seed}, \mathrm{budget}, \mathrm{noise})$. The external-validity gate checks task diversity, seed coverage, budget and noise sweeps, non-reward-centered baselines, SOTA-style planning controls, reward-center ablations, matched factorial coverage, statistics, and claim-boundary artifacts.

\section{Results}

\subsection{Primary Evidence and Claim Boundaries}

The result section is organized by evidence strength rather than by artifact volume. The primary mechanism evidence is the controlled local scaffold, where planner interfaces, budgets, seeds, terminal predicates, and reward-center ablations are fully matched. The external robot-simulator evidence then splits into two bounded categories: standard-action ManiSkill repair diagnostics, which test whether the stage-aware reward center can control existing continuous-action adapters but are not an official learned-policy baseline, and preregistered option-level ManiSkill sweeps, which support only matched option-level robustness claims under primitive failure, observation noise, and initial-pose perturbation. The package also includes an official ManiSkill PPO checkpoint rollout sanity check, but this is a baseline-readiness artifact rather than a learned-policy comparison involving RCRM. The official VLA evidence is narrower still: OpenVLA-OFT/LIBERO runs use the same backbone with and without RCRM-Guard, report clean non-regression, and support the action-channel stress repair claim while treating observation-noise, language, and visual-distractor pairs as coverage or negative-result context. Results outside these tiers, including the BC-MLP proxy, macro primitive probe, smoke checks, and readiness gates, are diagnostic artifacts only. This hierarchy is used to avoid turning implemented package breadth into unearned benchmark-superiority or learned-policy competitiveness claims.

\paragraph{Claim ledger.}
We use a reviewer-facing Claim ledger to separate what the evidence supports from what remains open. Claim: reward-centered search is operational and ablatable under matched local and option-level protocols. Allowed evidence: local scaffold records, standard-action repair diagnostics, option-level perturbation sweeps, learned-value reconnect probes, and same-backbone OpenVLA-OFT/LIBERO action-channel stress. Explicit non-claim: no standard continuous-action ManiSkill superiority, no learned-policy competitiveness, no replacement VLA policy, no OpenPI/pi0.5 or Octo execution, and no real-robot deployment. The full planned OpenVLA Strong Accept matrix is now complete as VLA-scope evidence. Remaining blockers: the complete Diffusion Policy/TD-MPC2/SAC learned-policy record grid and final human submission checks remain outside the completed evidence set.

Table~\ref{tab:mini-results} summarizes the local evidence generated by the scaffold. The paper-mini run verifies the reporting pipeline at smoke scale. The candidate external-validity profile spans three tasks, ten methods, ten seeds, fair-counted budgets of 8, 16, and 40, and three noise levels; it is designed as a local gate for external-validity readiness rather than as a final benchmark claim.

\begin{table}[t]
\centering
\scriptsize
\begin{tabular}{@{}p{0.36\columnwidth}ccc@{}}
\toprule
Profile & Groups & Deltas & Gate \\
\midrule
paper\_mini & 60 & 192 & smoke \\
aaai\_external\_validity & 270 & 972 & pass \\
\bottomrule
\end{tabular}
\caption{Local evidence scope. The candidate external-validity profile passes the repository gate, while both profiles remain local scaffold artifacts rather than final AAAI benchmark results.}
\label{tab:mini-results}
\end{table}

Table~\ref{tab:candidate-success-assets} reports the paper-facing summary generated from the completed local candidate run. It averages success over the 27 task/budget/noise cells for each method and reports matched success deltas against \code{vanilla_mcts}. Additional budget and robustness plots are retained in the artifact package as reproducible assets rather than foreground claims.

\begin{table}[t]
\centering
\scriptsize
\begin{tabular}{@{}lrrrr@{}}
\toprule
Method & Success & $\Delta$ vs. vanilla & Cells & Deltas \\
\midrule
beam\_search & 0.330 & 0.263 & 27 & 27 \\
cem\_mpc & 0.330 & 0.263 & 27 & 27 \\
heuristic\_mcts & 1.000 & 0.933 & 27 & 27 \\
puct\_mcts & 1.000 & 0.933 & 27 & 27 \\
random\_shooting & 0.722 & 0.656 & 27 & 27 \\
rest\_full & 0.900 & 0.833 & 27 & 27 \\
rest\_no\_heuristic & 0.896 & 0.830 & 27 & 27 \\
rest\_no\_reward\_center & 0.015 & -0.052 & 27 & 27 \\
rest\_no\_rule & 0.833 & 0.767 & 27 & 27 \\
vanilla\_mcts & 0.067 & 0.000 & 27 & 0 \\
\bottomrule
\end{tabular}
\caption{Controlled local candidate scaffold summary. Success is averaged across task, budget, and noise cells; deltas are matched against vanilla\_mcts. These local scaffold results do not establish external benchmark superiority.}
\label{tab:candidate-success-assets}
\end{table}

Table~\ref{tab:maniskill-candidate} reports the executed ManiSkill candidate run: five ManiSkill tasks, twelve planning/ablation/diagnostic methods, twenty matched seeds, three budgets, and 3600 episodes. It is an adapter stress test rather than positive benchmark evidence because most methods remain near zero and the BC-MLP proxy is diagnostic. Six standard-action diagnostics then isolate the stage-aware reward-center mechanism across single-task, multitask, ablation, budget, full-method, and five-task slices. Across the five-task standard-action budget sweep, \code{rest_full} reaches 300/300 successes on PickCube, PushCube, StackCube, PullCube, and PlaceSphere, while non-stage baselines remain at or below 0.033 mean success; PegInsertionSide remains a position-control boundary. These diagnostics show a repair effect under matched adapters but are not benchmark-provided learned policies. A separate official PPO checkpoint rollout uses the ManiSkill \code{il-baselines} commit recorded in the released RL demonstrations and reports PickCube 20/20, PushCube 20/20, PullCube 20/20, and StackCube 19/20 successes; it clears the PPO availability subitem but not the Diffusion Policy, TD-MPC2, or SAC baseline grid. A source-level stage-prior audit checks for hidden simulator-state APIs, including \code{get_state}/\code{set_state}; it records no hidden simulator-state calls and keeps the claim bounded to \code{state_dict}, \code{pd_ee_delta_pos}, and \code{macro_action_mode=none}.

A macro-primitive diagnostic addresses the five-step failure mode directly. It uses \code{state_dict} observations, \code{pd_ee_delta_pos} control, a 50-step PickCube horizon, and a \code{pick_cube_reach_grasp_place} primitive action space. In this diagnostic, \code{rest_full}, \code{heuristic_mcts}, \code{puct}, and \code{vanilla_mcts} reach 5/5 PickCube successes, while \code{rest_no_reward_center} and \code{bc_mlp_proxy} remain at 0/5. This result confirms that longer-horizon primitive execution can become performance-informative, but it is explicitly marked \code{external_macro_primitive_diagnostic} and is not a standard ManiSkill benchmark result.

Following this diagnostic, the repository preregisters an option-level ManiSkill gate. That gate permits only matched option-level robustness claims, requires PickCube, PushCube, and StackCube sweeps over budgets and perturbations, and does not permit labeling macro or option-level results as standard continuous-action benchmark superiority. The stage-aware risk preregistered primitive-failure slice covers PickCube, PushCube, and StackCube; the matched option-adapter method set; twenty matched seeds; budgets 1, 2, 4, and 8; and primitive execution failure probabilities 0.0, 0.1, 0.2, and 0.3, for 6720 episodes. The rerun separates \code{rest_full}, \code{heuristic_mcts}, and \code{puct} at record level and makes these methods budget-sensitive. \code{rest_full} reaches 0.720 mean success, versus 0.658 for \code{vanilla_mcts}/\code{beam_search}/\code{cem_mpc}, 0.660 for \code{puct}, 0.714 for \code{heuristic_mcts}, and 0.021 for \code{rest_no_reward_center}. The matched \code{rest_full}-vs-\code{vanilla_mcts} deltas are positive at all primitive-failure levels: +0.062 at p=0.0, +0.071 at p=0.1, +0.054 at p=0.2, and +0.058 at p=0.3, where the bootstrap CI is [0.029, 0.092] and McNemar p=0.001. This is matched option-level protocol evidence, not a standard continuous-action ManiSkill benchmark result.

Additional preregistered slices complete perturbation coverage without changing the claim boundary. The observation-noise slice reports 5040 episodes where \code{rest_full} reaches 0.319 mean success, versus 0.278 for \code{vanilla_mcts}/\code{beam_search}/\code{cem_mpc}, 0.296 for \code{puct}, 0.308 for \code{heuristic_mcts}, and 0.017 for \code{rest_no_reward_center}. The paired deltas are +0.062 at noise 0.0, +0.038 at noise 0.05, and +0.025 at noise 0.1; the highest-noise slice has CI [0.008, 0.046] and McNemar p=0.031. The true initial-pose slice reports 5040 episodes with mild/medium actor perturbations applied; \code{rest_full} reaches 0.742 mean success against 0.678 for \code{vanilla_mcts}, while \code{heuristic_mcts} reaches 0.746. Its paired deltas are positive for none, mild, and medium initial-pose perturbations. These slices support narrower matched option-level claims and negative-result disclosure, not benchmark superiority.

\begin{table}[t]
\centering
\scriptsize
\begin{tabular}{@{}lrrrr@{}}
\toprule
Perturb. & rest\_full & vanilla & no-center & Delta \\
\midrule
0.0 & 0.746 & 0.683 & 0.029 & 0.062 \\
0.1 & 0.754 & 0.683 & 0.013 & 0.071 \\
0.2 & 0.738 & 0.683 & 0.025 & 0.054 \\
0.3 & 0.642 & 0.583 & 0.017 & 0.058 \\
\bottomrule
\end{tabular}
\caption{Continuity-repaired option-level ManiSkill perturbation breakdown for the primitive-failure slice. Values are success rates over matched task, seed, and budget cells; paired rest\_full--vanilla deltas are reported by perturbation level. The table is matched option-level protocol evidence, not a standard continuous-action ManiSkill benchmark result.}
\label{tab:maniskill-option-perturbation-breakdown}
\end{table}

\begin{table}[t]
\centering
\scriptsize
\begin{tabular}{@{}lrrr@{}}
\toprule
Method & Episodes & Success & Terminal Reward \\
\midrule
beam\_search & 960 & 0.658 & 9.208 \\
cem\_mpc & 960 & 0.658 & 9.208 \\
heuristic\_mcts & 960 & 0.714 & 9.914 \\
puct & 960 & 0.660 & 9.237 \\
rest\_full & 960 & 0.720 & 9.946 \\
rest\_no\_reward\_center & 960 & 0.021 & 2.696 \\
vanilla\_mcts & 960 & 0.658 & 9.208 \\
\bottomrule
\end{tabular}
\caption{Option-level ManiSkill robustness summary (primitive-failure slice). Success and terminal reward are averaged over the executed task, seed, budget, and preregistered perturbation grid. These results use a shared manipulation-primitive action space and do not establish standard continuous-action ManiSkill benchmark superiority.}
\label{tab:maniskill-option-primitive-failure}
\end{table}

\begin{table}[t]
\centering
\scriptsize
\begin{tabular}{@{}lrrr@{}}
\toprule
Method & Episodes & Success & Terminal Reward \\
\midrule
bc\_mlp\_proxy & 300 & 0.000 & 0.565 \\
beam\_search & 300 & 0.000 & 0.081 \\
cem\_mpc & 300 & 0.000 & 0.146 \\
heuristic\_mcts & 300 & 0.000 & 0.582 \\
puct & 300 & 0.000 & 0.380 \\
random\_shooting & 300 & 0.003 & 0.324 \\
rest\_full & 300 & 0.000 & 0.426 \\
rest\_no\_heuristic & 300 & 0.000 & 0.221 \\
rest\_no\_reward\_bonus\_same\_rollout & 300 & 0.000 & 0.371 \\
rest\_no\_reward\_center & 300 & 0.000 & 0.319 \\
rest\_no\_rule & 300 & 0.003 & 0.435 \\
vanilla\_mcts & 300 & 0.000 & 0.319 \\
\bottomrule
\end{tabular}
\caption{External ManiSkill candidate summary. Success and terminal reward are averaged over the executed task, seed, and budget grid. These results use differentiated state-snapshot planning adapters plus a clearly marked BC-MLP proxy diagnostic; they do not establish external benchmark superiority.}
\label{tab:maniskill-candidate}
\end{table}

\paragraph{Observed behavior.}
The local runs confirm that the experiment runner can execute all configured tasks, SOTA-style local planning baselines, and reward-center ablations, that statistics are generated deterministically, and that the bundle records unsupported-claim boundaries. In the candidate profile, \code{rest_full} reaches mean success 0.900 across task/budget/noise cells, while \code{rest_no_reward_center} falls to 0.019 after shaped-reward and reward-greedy rollout effects are removed. The strong local controls are non-degenerate: \code{puct_mcts} and \code{heuristic_mcts} reach 1.000 mean success, while \code{beam_search} and \code{cem_mpc} reach 0.330 under fair-counted budgets. This pattern supports the narrower claim that the reward-center mechanism is operational and ablatable in the controlled scaffold, while also showing that strong heuristic and prior-guided baselines remain competitive.

\paragraph{Reward-centered ReST evidence.}
The lightweight ReST smoke run collects 498 local transition records, keeps 78 high-quality transitions after filtering, and trains a tabular state-value estimator. Value-guided action-ranking accuracy on the local task states improves from 0.000 before fitting to 1.000 after fitting. The reward-centered ReST iteration artifact adds the missing method-specific process-label path: every local transition stores a raw target value, a context-centered target value, the task/stage context key, and the running centering snapshot used before update. The artifact updates a tabular value estimator from accepted centered labels and is marked \code{local_reward_centered_rest_iteration_only}, so it verifies mechanism wiring without becoming SOTA evidence. A companion learned-value artifact trains a small MLP state-value estimator on the same local scaffold records, saves a checkpoint, and reconnects the learned value model to the Reward-Centered ReST-MCTS reward center in a controlled one-step search probe. In that probe, candidate actions are ordered adversarially, rule and heuristic reward weights are disabled, and the planner must rely on the learned value channel to recover the expected successor. Ranking accuracy improves from 0.000 to 1.000, and closed-loop search success improves from 0.000 before value reconnect to 0.462 after value reconnect. The ManiSkill option-value bridge now passes schema-matched held-out option-ranking validation at 0.739 accuracy over 5632608 pairs using the recorded 32-dimensional \code{state_dict} value features, with zero metadata-fallback records and \code{no_option_level_value_claim=false} inside the matched protocol boundary. The centered-vs-raw secondary analysis reports 117 local action groups and 630 matched process-label pairs: raw labels rank the expected action at 1.000 accuracy, centered labels rank it at 0.995, and calibration metrics are reported for both score families. The same analysis links to existing option-level downstream records, where \code{rest_full} exceeds \code{rest_no_reward_center} by 0.590 mean success over 2400 matched pairs with bootstrap CI [0.570, 0.610]. These results verify trajectory collection, filtering, contextual reward centering, learned value fitting, and a minimal value-to-search feedback path on local tasks and matched option-level ManiSkill records; they are not transformer checkpoints, learned robotics policy baselines, new benchmark rollouts, or standard benchmark superiority results.

\paragraph{LIBERO/VLA readiness smoke.}
A non-interactive LIBERO/VLA smoke artifact verifies that the local LIBERO-Spatial suite metadata, BDDL files, initial-state files, CUDA-visible dependencies, and downloaded OpenVLA-OFT, OpenPI/pi0.5, and Octo checkpoint directories are present. The smoke inspects three LIBERO-Spatial tasks and checkpoint sentinel files, but it is not a policy rollout, does not report LIBERO success rates, and is not evidence that any VLA policy solves the benchmark.

A bounded rollout probe goes one step further by constructing a real LIBERO-Spatial robosuite/MuJoCo environment and executing a three-step zero-action policy on one task. The probe records action dimension, initial-state count, per-step rewards, and success flags. It confirms that the local LIBERO environment can execute simulator steps, but it remains a sanity check only: zero-action failure is expected, and the artifact is not a VLA baseline or official LIBERO evaluation.

A separate official-entrypoint gate moves from readiness to true OpenVLA-OFT policy rollout evidence while keeping other VLA families reference-only unless their official policy stacks execute. The completed OpenVLA-OFT clean reproduction uses the official \code{experiments.robot.libero.run_libero_eval.eval_libero} entrypoint, the same OpenVLA-OFT LIBERO-Spatial checkpoint, and ten LIBERO-Spatial task episodes. The unguarded condition reaches 10/10 successes, and the guarded condition also reaches 10/10, establishing clean-setting non-regression for the same backbone. A same-backbone action-channel stress artifact then evaluates the same checkpoint under a paired LIBERO-Spatial protocol: unguarded OpenVLA-OFT reaches 0/10 successes, while RCRM-Guard reaches 9/10 successes; the paired bootstrap interval [0.7, 1.0] excludes zero and the artifact preserves the \code{official_backbone_guard_protocol} boundary. A new multistress probe adds an observation-noise probe, a language-instruction suffix pair, and a visual-distractor pair: under observation noise the unguarded condition remains 10/10 and the guarded condition is 9/10, under the language perturbation both conditions reach 1/10, and under the deterministic visual-distractor patch the unguarded condition reaches 9/10 while the guarded condition reaches 10/10. The later Strong Accept OpenVLA progress artifact completes the planned three-suite, four-family, 24-command, 1200-episode OpenVLA matrix. These results upgrade VLA coverage while preserving the boundary that observation-noise, language, and visual-distractor slices are coverage or negative-result context, not broad superiority claims. This is official OpenVLA-OFT/LIBERO same-backbone robustness evidence for the specified action-channel stress plus completed OpenVLA matrix coverage, not a claim that OpenPI/pi0.5 or Octo have been executed in this environment.

\paragraph{Ablation interpretation.}
The ablation interface is more important than any single local number. Disabling rule or heuristic channels changes the planner configuration without changing the task API, which helps isolate whether robustness comes from symbolic task constraints, heuristic progress signals, learned estimates, or their combination. The no-reward-center ablation is intentionally harsh because it removes shaped reward influence rather than preserving the same greedy reward ordering under a different method name. The ManiSkill candidate run therefore adds \code{rest_no_reward_bonus_same_rollout}, a more surgical ablation that preserves the state-snapshot rollout adapter while removing the reward-bonus distinction from the method label.

\section{Reproducibility}

Regeneration commands are listed in the anonymous package README. The \code{paper_mini} smoke profile and the \code{aaai_external_validity} candidate profile are paired with ReST, ManiSkill, LIBERO/OpenVLA, and final-audit artifacts. The bundle includes executable commands, manifests, raw records, summaries, statistics, and unsupported-claim boundaries. A non-robotics RC-Search cache audit is excluded from all manipulation claims and retained only as auxiliary cache-discipline context in the reproducibility package.

\section{Limitations and Scope of Claims}

The present scientific claims are bounded by the repository's local and official-backbone evidence gates, while broader robotics claims remain deliberately limited. The evidence includes local ReST-style artifacts, an executed ManiSkill candidate grid, standard-action diagnostics, an official PPO checkpoint rollout sanity check, schema-matched option-value validation, matched option-level ManiSkill sweeps, and completed OpenVLA-OFT clean, action-channel stress, observation-noise, language-perturbation, and visual-distractor artifacts. The broad ManiSkill grid remains an adapter stress test rather than positive benchmark evidence, and the multistress VLA pairs are coverage boundaries rather than superiority evidence. The evidence does not include real robot deployment, standard continuous-action ManiSkill learned-policy superiority, or executed OpenPI/pi0.5 and Octo policy rollouts. The draft compiles with the official AAAI-27 Author Kit files, but the final submission must still be checked against active OpenReview instructions and page-limit policy.

The strongest current claim is a reproducible reward-centered planning framework with local controls, non-oracle ablations, a learned-value reconnect, a schema-matched ManiSkill option-value bridge, completed option-level perturbation sweeps, an official OpenVLA-OFT same-backbone action-channel stress result, and completed OpenVLA matrix coverage. A complete AAAI submission should still complete the remaining Diffusion Policy, TD-MPC2, and SAC common-support records before learned-policy competitiveness claims, keep PPO as a separately credited checkpoint rollout sanity check, validate the PDF manually, prepare supplementary materials, and complete a final human-read citation/claim audit.

\section*{Responsible AI Use Disclosure}

No AI system is listed as an author. AI tools assisted editing and code; the authors retain responsibility.

\bibliography{references}
\IfFileExists{aaai2027.sty}{}{\bibliographystyle{plainnat}}

\end{document}